\documentclass[runningheads]{llncs}

\usepackage{eccv}

\usepackage{eccvabbrv}

\usepackage{graphicx}
\usepackage{booktabs}

\usepackage[accsupp]{axessibility}  

\usepackage[pagebackref,breaklinks,colorlinks,citecolor=eccvblue]{hyperref}

\usepackage{orcidlink}

\begin{document}

\title{Adaptive Multi-Modal Control of Digital Human Hand Synthesis Using a Region-Aware Cycle Loss} 

\titlerunning{Multi-Modal Control of Digital Human Hand Synthesis}

\author{Qifan Fu\inst{1}\orcidlink{0000-0003-3505-9865} \and
Xiaohang Yang\inst{1}\orcidlink{0000-0002-8103-5664} \and
Muhammad Asad\inst{1}\orcidlink{0000-0002-3672-2414} \and
Changjae Oh\inst{1}\orcidlink{0000-0002-6522-2451} \and
Shanxin Yuan\inst{1}\orcidlink{0000-0002-6918-8588} \and
Gregory Slabaugh\inst{1}\orcidlink{0000-0003-4060-5226}}

\authorrunning{Q.Fu et al.}

\institute{Queen Mary University of London, London, UK \\ 
\email{q.fu@qmul.ac.uk}}

\maketitle

\begin{abstract}
Diffusion models have shown their remarkable ability to synthesize images, including the generation of humans in specific poses. However, current models face challenges in adequately expressing conditional control for detailed hand pose generation, leading to significant distortion in the hand regions. 
To tackle this problem, we first curate the How2Sign dataset to provide richer and more accurate hand pose annotations. In addition, we introduce adaptive, multi-modal fusion to integrate characters' physical features expressed in different modalities such as skeleton, depth, and surface normal. Furthermore, we propose a novel Region-Aware Cycle Loss (RACL) that enables the diffusion model training to focus on improving the hand region, resulting in improved quality of generated hand gestures. More specifically, the proposed RACL computes a weighted keypoint distance between the full-body pose keypoints from the generated image and the ground truth, to generate higher-quality hand poses while balancing overall pose accuracy. Moreover, we use two hand region metrics, named \emph{hand-PSNR} and \emph{hand-Distance} for hand pose generation evaluations. Our experimental evaluations demonstrate the effectiveness of our proposed approach in improving the quality of digital human pose generation using diffusion models, especially the quality of the hand region. 
The source code is available at \url{https://github.com/fuqifan/Region-Aware-Cycle-Loss}.
\end{abstract}

\section{Introduction}
\label{sec:intro}

\begin{figure}[tb]
  \centering
  \includegraphics[height=6.5cm]{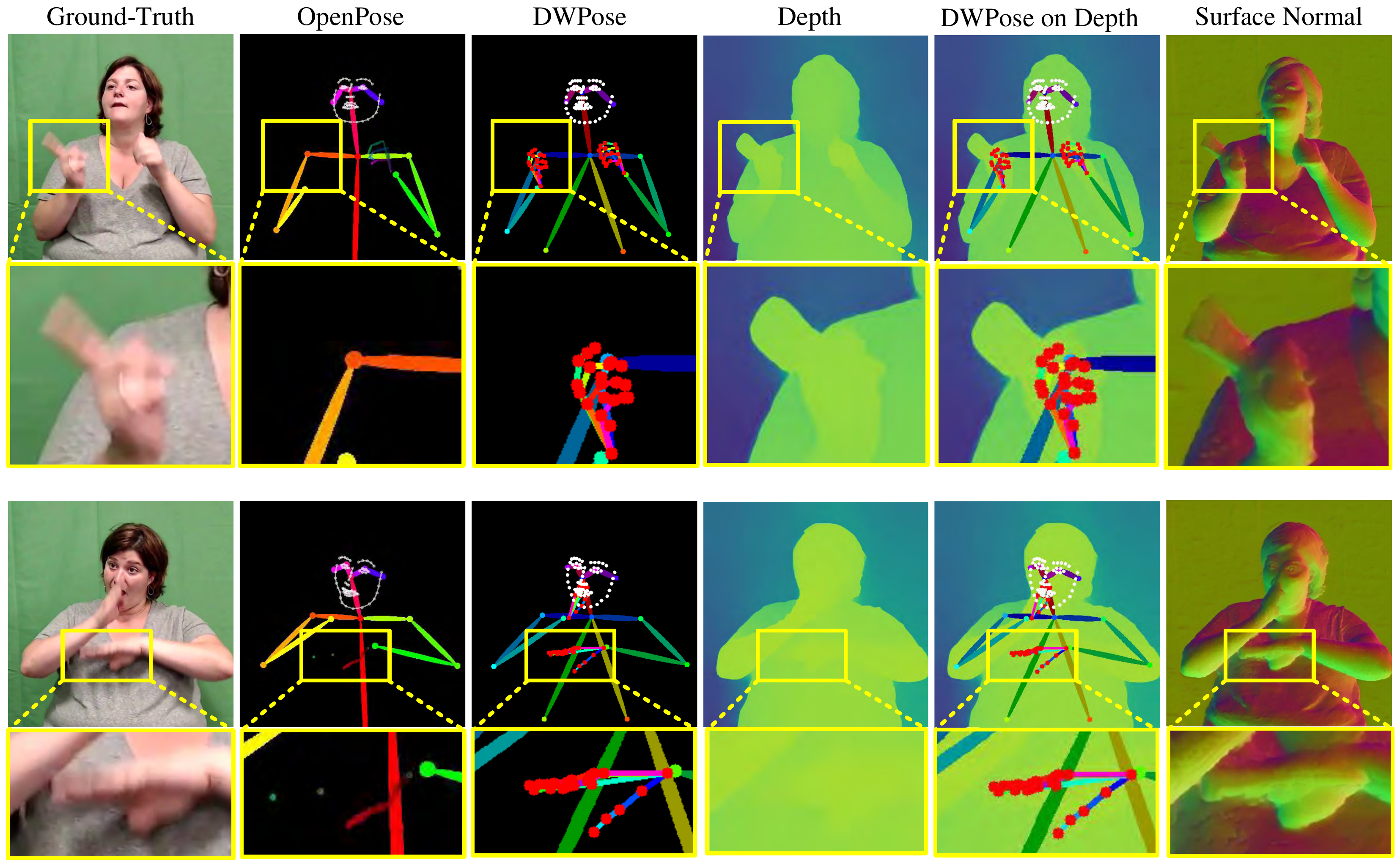}
      \caption{Annotation of human poses with different methods. The recently introduced DWPose method depicts higher quality keypoints and skeleton annotations as compared to OpenPose. We note that the skeleton lacks finer details such as the person's body shape. Depth annotation provides additional body shape information, however, when the hand moves near the body, it is difficult to distinguish hands in the depth map. In contrast, the surface normal provides a more detailed label capturing both the body shape, skin texture, as well as the surface texture of clothes. But the surface normal is sensitive to motion blur. Particularly, the outputs of different annotated modalities may conflict (marked with yellow boxes), which frequently occurs with the skeleton and depth map.
  }
  \label{fig:Fig1}
\end{figure}

In recent years, there has been remarkable progress in generative AI for images~\cite{ramesh2022hierarchical, avrahami2023chosen, pham2024cross} and videos~\cite{wang2024magicvideo, arkhipkin2023fusionframes, guo2023sparsectrl, ren2024consisti2v}. Building on these achievements, digital human synthesis, an indispensable element in many digital environments, has gathered increased attention~\cite{feng2023dreamoving, hu2023animate, xu2024you, shi2024motion}.
Existing digital human pose generation models can generate a body pose consistent with an input skeleton, however, these methods often fail to generate complex and accurate hand gestures. There are two reasons for this problem; the first is due to the complex spatial structure and complicated motions of the hands, resulting in a scarcity of high-quality hand annotations as shown in Figure \ref{fig:Fig1}, and the second is due to the lack of training methods and evaluation metrics targeted for hand features. 
Although the information density of the hand regions in an image is much higher than that of the rest of the body, few existing generative methods have been able to make the model pay attention to this difference.
To address this gap, it is necessary to choose scenes with a large proportion of hands, such as sign language, to construct a dataset and design training strategies to provide additional guidance focusing on the hands while learning the body pose.

In this work, our goal is to make pose-controllable digital human generation with high-quality complex hand poses.
Towards this goal, we choose the How2Sign dataset~\cite{duarte2021how2sign}, a large-scale sign language dataset comprised of videos with rich and complex hand gestures. We chose this dataset because sign language is an important exemplar of upper body movement and contains complex and semantically rich hand gestures and facial expressions. Progress on this dataset can enable downstream tasks related to upper body postures. We improve the quality of the How2Sign dataset by identifying high quality, non-blurry frames to train and expand the annotation with additional modalities for conditional control of the image generation. As shown in Figure \ref{fig:Fig1}, different annotated modalities have different strengths and weaknesses, and some may conflict with each other. To make the model learn complementary features from multi-modal control information and inspired by~\cite{liu2023hyperhuman}, we design an adaptive multi-modal control feature fusion network for trainable ControlNet~\cite{zhang2023adding} and fixed pre-trained Stable Diffusion XL~\cite{podell2023sdxl}. Different from the method of ~\cite{liu2023hyperhuman}, we predict the fusion weights of the different modalities based on the inputs, by training an adaptive weight prediction network.

To further enhance the accuracy of complex hand regions, we propose a Region-Aware Cycle Loss (RACL). In particular, the RACL is computed based on a weighted sum of the Euclidean distance between the keypoints of the generated images and the keypoints of the ground truth images. By designing the weights of different regions, we make the model improve the quality of hand region generation while ensuring the quality of overall pose generation.

We utilize our improved dataset with multi-modal annotations and our proposed RACL to train the ControlNet model, which when paired with the diffusion model, results in improved pose alignment of synthesized digital human generation with ground truth detailed hand pose. 
Specifically, we experiment with different modalities (depth, skeleton, surface normal, as shown in \cref{fig:Fig1}) for controlling digital human generation, use two hand region metrics, named hand-PSNR and hand-Distance, respectively, to evaluate the generation quality of the hand region, and observe the surface normal produces superior results for single modality control of hand poses. Furthermore, we found that the existing widely used skeletal annotation and depth annotation modalities fail to provide accurate hand pose information. In contrast, the surface normal not only provides more accurate pose and shape information resulting in improved hand pose and facial expression generation, but it also provides surface texture information, which improves the quality of generated skin and hair. In addition, our proposed RACL with weights for different regions can guide the model to focus more on learning hand features and further improve the accuracy of hand pose.

The contributions of this work are three-fold:
\begin{enumerate}
    \item We address the task of high-quality hand gesture generation for digital humans, and devise novel strategies for dataset filtering and conditional control of hand pose based on different inputs (depth, skeleton, and surface normal). 
    \item We design an adaptive multi-modal control fusion network, which can enhance the model's ability to learn from complex input modalities for better pose alignment within digital human generation. Especially when there are conflicts between different control modalities.
    \item We design a Region-Aware Cycle Loss (RACL), which can enhance the model's ability to learn complex hand gestures and improve the generation accuracy of hand gestures, while ensuring quality of upper body posture. 
\end{enumerate}

\section{Related Work}
Our work is related to pose-controlled digital human video generation, multi-modal control fusion for generative AI and photo-realistic Sign Language Production (SLP). Recent literature is summarized below, along with the similarities and differences of these existing methods compared to our proposed approach.

\subsection{Pose-Controlled Digital Human Generation}
In this paper, ``pose-controlled'' refers to the control over the pose of the person in generated image, but it is not limited to using only keypoint skeletons for control. Earlier work on digital human generation utilized Generative Adversarial Networks (GANs)~\cite{chan2019everybody, liu2021liquid}. However, GANs have limitations including instability during training and mode collapse, rendering them more difficult to control than diffusion models. Recently, diffusion models have superseded GANs in the field of pose-controlled digital human generation~\cite{hu2023animate, feng2023dreamoving, sn_handiffuser_cvpr_2024}. Animate Anyone~\cite{hu2023animate} proposed a pipeline customised for character animation based on the diffusion model with a novel ReferenceNet. 
Although Animate Anyone can generate action videos of arbitrary characters, it doesn't perform well when generating complex hand gestures. 
HanDiffuser~\cite{sn_handiffuser_cvpr_2024} used SMPL-Body and MANO-Hand parameters to generate images with realistic hands. But 3D parameters that provide a rich source of information are not easily available. For example, we tried to annotate the How2Sign dataset with 3D parameters using the latest OSX~\cite{lin2023one} model, but the accuracy of the 3D parameters obtained is lower than that of the surface normal obtained using Omnidata~\cite{eftekhar2021omnidata} annotation.
Another work, DreaMoving~\cite{feng2023dreamoving}, introduces skeleton and depth control to make the generated human poses more controllable. Depth information supplements the missing character shape information of the skeleton, but lacks detailed hand and facial features. In addition, the depth information also lacks texture detail, which is noticeable in the details of the hair. In contrast to the focus of the aforementioned work, our method focuses on detailed hand gesture-controllable digital humans. To this end, we explore the effect of different control information and design an adaptive fusion network for multi-control of digital human gestures. We also propose a new loss, named RACL, to adjust the model's attention to different parts of the body, allowing the model to learn hand features in a more focused way while balancing the quality of the generation of other parts of body posture.

\subsection{Multi-modal Control Fusion for Generative AI}
Recent literature has attempted the fusion of different control modalities to compensate for the shortcomings of a single modality and to enhance the generation results~\cite{liu2023hyperhuman, zhu2024champ, Lin2024CtrlAdapter}. Hyperhuman~\cite{liu2023hyperhuman} attempted the fusion of multiple control modalities to improve the quality of model generation, but did not observe any conflicts between the different modalities. Champ~\cite{zhu2024champ} improves the generation quality by fusing multiple control modalities from 3D parameters with the 2D skeleton, 
but does not mention and provide a solution for the possible conflicts of multiple modalities. The conflicts of different modalities is especially serious in the hand-pose related generation, so we designed an adaptive weight prediction network to predict the fusion weights of different modalities based on different inputs, which allows the model to take advantage of the strengths from different modalities. CtrlAdapter~\cite{Lin2024CtrlAdapter} provides an interface-rich adaptive fusion module that enhances multi-control fusion for image and video generation. However, CtrlAdapter requires multiple ControlNets pre-trained on different control modalities. In contrast, our pipeline only trains lightweight encoders for different modalities and uses a shared ControlNet to encode all modalities.

\subsection{Photo-Realistic Sign Language Production}
Sign language is an ideal application for upper body movement synthesis, with important societal impact, and has received notable attention in the literature. However, traditional approaches to SLP focus on the generation of skeletal postures~\cite{saunders2020progressive, saunders2021continuous, xie2022vector}. Recently, photo-realistic SLP methods~\cite{saunders2020everybody, saunders2022signing} have started to appear. SignGAN~\cite{saunders2020everybody} is the first SLP framework to produce photo-realistic sign language videos directly from spoken language, catering to the demand for systems that are comprehensible and accepted by deaf communities. It employs a transformer architecture to convert spoken language into skeletal pose sequences and a pose-driven human synthesis model for generating photo-realistic videos from these poses, improving the fidelity of sign language creation. The framework includes an innovative keypoint-centered loss function, named the hand keypoint loss, to enhance the quality of hand images and supports controllable video production, showing exceptional effectiveness over traditional approaches in both quantitative assessments and human perceptual evaluations. Subsequently, a novel Frame Selection Network (FS-Net)~\cite{saunders2022signing} is proposed to improve temporal alignment in coarticulated signing, along with SignGAN, with a pose-conditioned human synthesis model to create photo-realistic sign language videos from skeleton poses, enhanced with a keypoint-based loss function for better hand image quality. These approaches have shown improvements in coarticulation, realism, and comprehension by native deaf signers over baseline methods. However, the hand keypoint loss of SignGAN uses an extra discriminator to distinguish good and bad quality of the generated hand region keypoints. This GAN-based strategy is not easy to train and does not directly reflect the generation quality of the current sample. And emphasizing only the hand region by the hand keypoint loss may cause the model to overfit and ignore other body parts. Different from the above GAN-based approach, our approach generates higher-quality digital humans based on a more powerful and easier to train diffusion model with higher scalability. Meanwhile, our discriminator-free RACL provides more efficient and direct feedback on the quality of the generated pose. Moreover, the weight parameters can be adjusted so that the model can be trained to pay more attention to the hand region while taking into account the learning of other body parts to avoid overfitting the hand region.

\section{Method}
\subsection{Preliminaries}
The diffusion model operates as a generative model that introduces additive noise to data in a forward process, while in a reverse process, it predicts and subtracts the corresponding noise from the data. Our method utilizes the Stable Diffusion XL model, a Latent Diffusion Model (LDM)~\cite{rombach2022high}, as the foundational backbone of our framework. In contrast to pixel-level denoising, LDMs conduct denoising within the low-dimensional latent space of a variational autoencoder (VAE)~\cite{kingma2013auto}. This approach facilitates efficient training and inference. In the forward process of noise addition, the VAE encoder extracts the image features \({F_0}\) from the input image ${I_0}$. Subsequently, Gaussian noise $\varepsilon  \sim {\cal N}\left( {0,{\rm{ }}1} \right)$ is gradually added to \({F_0}\) to obtain a noised feature representation, as:
\begin{equation}
  {F_t}{\rm{ }} = {\rm{ }}{\alpha _t}{F_0}{\rm{ }} + {\rm{ }}{\sigma _t}\varepsilon ,
  \label{eq:1}
\end{equation}
where ${\alpha _t},{\rm{ }}{\sigma _t}{\rm{ }} \in {\rm{ }}\left[ {0,{\rm{ }}1} \right]$ are two predefined functions of time step $t \sim {\cal U}\left( {1,{\rm{ }}.{\rm{ }}.{\rm{ }}.{\rm{ }},{\rm{ }}T} \right)$, satisfying the principle of variance-preservation, i.e.,  $\alpha _t^2 + \sigma _t^2 = {\rm{ }}1.$ The diffusion model ${D_\theta }$  learns the reverse denoising process to obtain noise-free features after a set of iterations. The conditional diffusion model ${D_\theta }$  is trained by the following MSE loss:
\begin{equation}
  {{\cal L}_\varepsilon } = {\mathbb E_{F_0^{1:T},c,\varepsilon ,t}}\left[ {\left\| {\varepsilon  - {D_\theta }\left( {{\alpha _t}F_0^{1:T}{\rm{ }} + {\rm{ }}{\sigma _t}\varepsilon ,c,t} \right)} \right\|_2^2} \right],
  \label{eq:2}
\end{equation}
where $c$ represents the embeddings of any condition e.g., text, image, and audio.

\subsection{Dataset Curation}
\begin{figure}[tb]
  \centering
  \includegraphics[height=5.7cm]{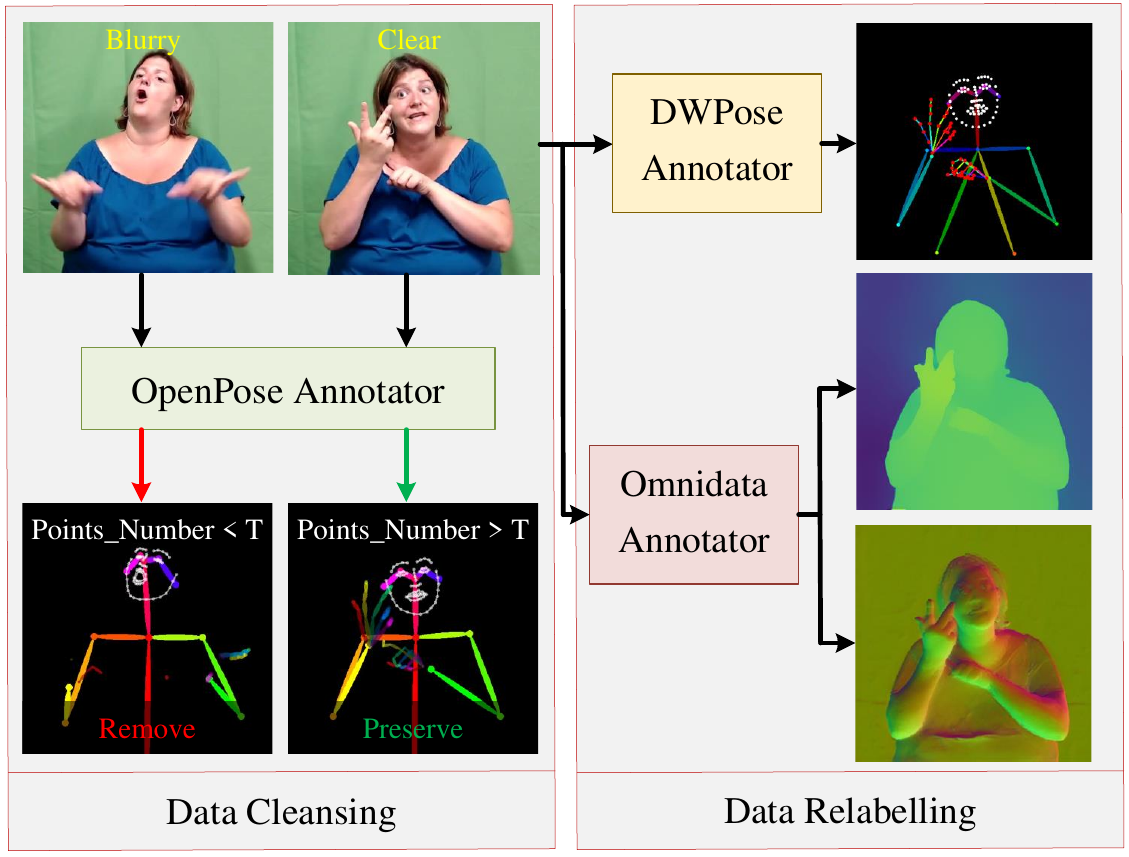}
      \caption{Our data pre-processing pipeline consists of two steps, data cleansing and data relabelling. In the data cleansing stage, based on the prediction of the OpenPose annotator, the frames of motion-blurred hands are filtered, leaving clear frames for the second stage of annotation. Then in the data annotation stage, the latest DWPose and Omnidata annotators are used to annotate the clear frames to get the corresponding modality annotations.
  }
  \label{fig:Fig2}
\end{figure}

Commonly utilized human pose annotation tools, such as OpenPose~\cite{8765346}, Residual Steps Network (RSN)~\cite{cai2020learning}, and Distribution-Aware coordinate Representation of Keypoint (DARK) pose~\cite{zhang2020distribution}, excel in annotating keypoints and character skeletons within images or videos. Nevertheless, the motion blur present in recorded videos can lead to blurred hands in some RGB frames. As illustrated in Figure \ref{fig:Fig1}, OpenPose annotation tools may lack fine-grained annotation for the complete human body pose, particularly for hands, and may struggle with fast or complex movements. Recognizing the importance of high-quality training data, characterized by clear RGB frames and precise annotations, especially for hands, we propose a two-stage dataset pre-processing pipeline outlined in Figure \ref{fig:Fig2} to ensure optimal conditions for the conditional control of the generated digital human pose.

\subsubsection{Data Cleansing}

The dataset pre-processing pipeline begins with data cleansing. We use OpenPose to detect motion blur in RGB images and leave the clean images without severe motion blur for labelling. Specifically, frames for which OpenPose fails to detect the hand keypoints are discarded. These frames typically have substantial motion blur in the hand regions. Considering that some missing keypoints come from occlusion of the face or torso, we determine the criticality of missing keypoints by setting a threshold on the overall number of keypoints,  i.e. whether the number $n$ of whole-body keypoints ${K^f} = \left( {\kappa _1^f,\kappa _2^f, \cdots ,\kappa _n^f} \right)$ obtained by OpenPose for a frame $f$ are less than a certain preset threshold $\tau $ to determine whether the whole-body keypoints are sufficiently complete in general. When counting the number of whole-body keypoints, we consider the confidence level of the hand keypoints predicted by the OpenPose model, and only hand keypoints with a sufficiently high confidence level are counted. Then we discard frames with severely missing keypoints and pass the filtered clear frames to the second stage for relabelling.

\subsubsection{Data Relabelling}
In the second stage, we conduct a re-annotation of the filtered frames using the DWPose~\cite{yang2023effective} and Omnidata~\cite{eftekhar2021omnidata} annotators to obtain the skeleton, depth and surface normal annotations. As illustrated in Figure \ref{fig:Fig1}, although DWPose yields better annotations than OpenPose, the skeleton and keypoint annotations lack information about personalised features. This limitation becomes especially apparent in elements like a character's hair and clothing, which are entirely overlooked in the skeleton annotations. To address this gap, we employ the Omnidata annotator for the filtered clear frames to obtain depth and surface normal annotations. These depth and normal annotations enrich the dataset with more comprehensive character and scene information, while the depth map fails to distinguish fingers. Especially when the hand is close to the chest, the depth map cannot distinguish between the hand and the chest. To make skeleton and depth complementary, we draw the skeleton on depth, which gives us the annotation modality DWPoseonDepth, as depicted in Figure \ref{fig:Fig1}. However, this brings out the conflict between the keypoint skeleton and the depth, especially in the hand area. For example, as shown the yellow boxes in the first row in Fig. 1, the skeleton is not labeled with the outstretched right hand fingers, whereas both depth and surface normal are labeled with the outstretched fingers. Notably, the surface normal annotation provides extensive details about a character's posture, physique, and the texture of hair and skin. But the surface normal annotation is sensitive to motion blur. Considering the conflicts that exist in different modalities, as well as the advantages and disadvantages of the various modalities, simply expanding the dataset is not enough, and fusion networks must be designed to complement the strengths and weaknesses of the various modalities.

\begin{figure}[tb]
  \centering
  \includegraphics[height=5.5cm]{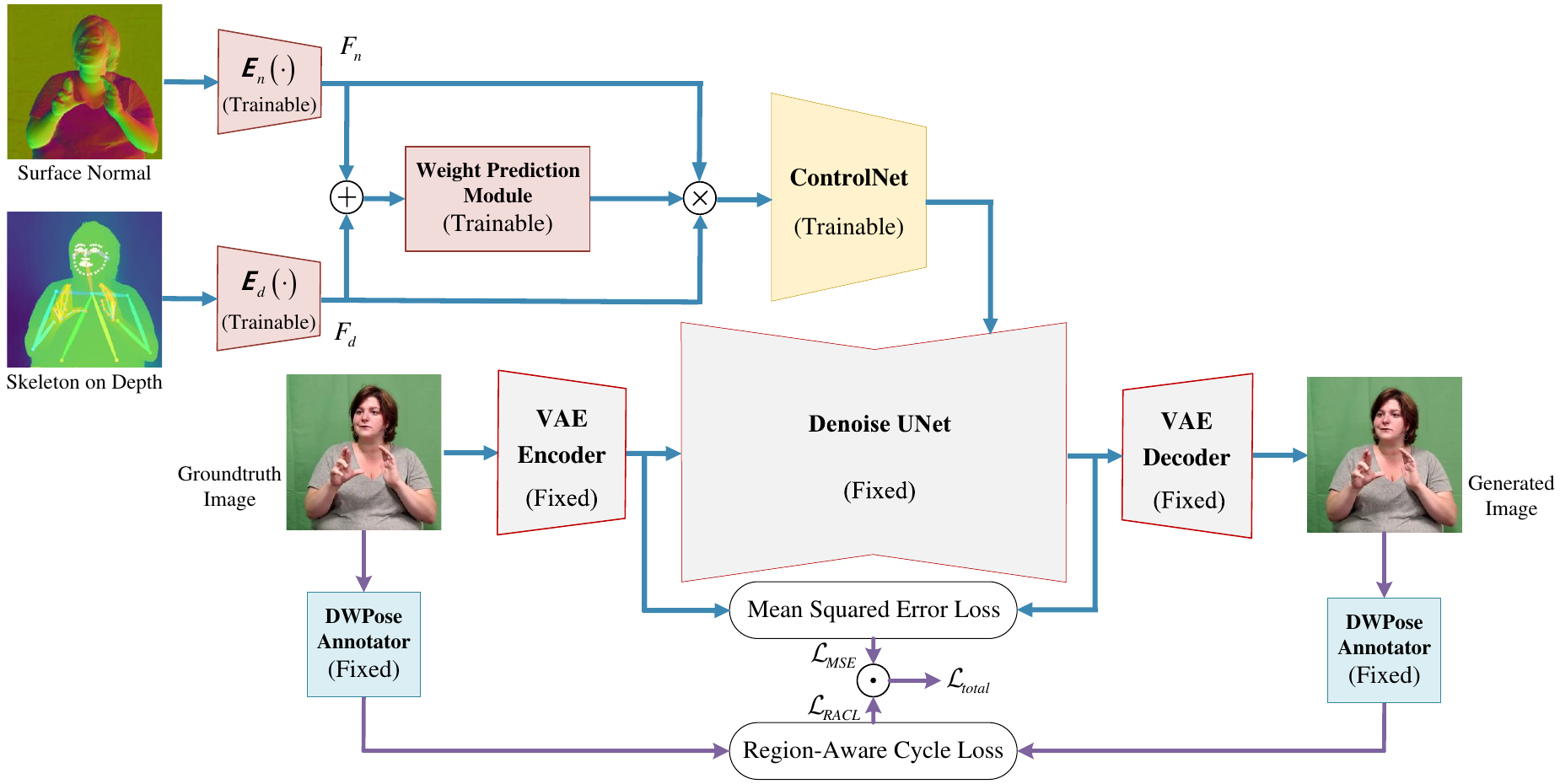}
      \caption{The model training pipeline with the adaptive multi-control fusion network and Region-Aware Cycle Loss (RACL). The weight prediction module predicts the adaptive
weights based on different control features input, and the  features for fusion are obtained by
weighted summation of these features. We use DWPose to obtain the keypoint coordinates of the generated image and the ground truth frame. Then the  Euclidean distance for these keypoints is calculated, weighted and summed as the RACL. The combination of the RACL and MSE loss enhances the learning of hand features in the model training. Note, at inference, no ground truth data are input.
  }
  \label{fig:Fig3}
\end{figure}

\subsection{Adaptive Multi-modal Control Fusion Network for ControlNet}
To address the conflicting conditions of the different modalities identified during the dataset curation, and to enable multiple modalities to complement each other, we designed an adaptive approach to fuse the multiple control modalities. Figure \ref{fig:Fig3} shows the proposed details of the adaptive multi-modal control fusion network for ControlNet. The fusion network is suitable for fusing with more control information inputs, but we only show the fusion of the two control input, which performs best in our experiments in this paper. Specifically, we use two tiny networks ${E_{d}}\left( \cdot \right)$ and ${E_{n}}\left( \cdot \right)$, each of which has four convolution layers with $4 \times 4$ kernels and $2 \times 2$ strides,  to encode corresponding control inputs into feature maps $F_d$ and $F_n$.

Then, the concatenated feature maps of $F_d$ and $F_n$ are put into a weight prediction network, which consists of 8 convolution layers to predict the Normalised spatial weights for each region of each mode. Subsequently, the predicted weights are used to weighted sum the corresponding region of the input control features to obtain the fused features. Next, the ControlNet encodes the fused features into different scale feature vectors. Finally, these feature vectors are fed into the corresponding layers of UNet as control information for denoising.

\subsection{Region-Aware Cycle Loss (RACL) for Model Training}
Diffusion models are typically trained using MSE loss ${{\cal L}_\varepsilon }$ in Eq. \ref{eq:2}, which mainly impacts overall denoising, resulting in high-quality image generation. A challenge when applying these methods on digital human generation is that it does not directly provide relevant human pose guidance during training and hence results in low quality human pose generation. Towards this end, we propose Region-Aware Cycle Loss (RACL) which aims to improve the model's ability to learn and generate high-quality digital humans, especially hand poses. Figure \ref{fig:Fig3} illustrates the model training pipeline using RACL. Specifically, we employ the DWPose tool to annotate hands, face and body keypoints for the images generated by our proposed model as well as for the original ground truth images. 
Our proposed RACL is defined as follows:
\begin{equation}
{{\cal L}_{RACL}} = {\beta _{hands}} \cdot {{d}_{hands}} + {\beta _{face}} \cdot {{d}_{face}} + {\beta _{body}} \cdot {{d}_{body}},
  \label{eq:3}
\end{equation}
where ${d}_{hands} = \sum\limits_{j = 0}^{J} {\sqrt {\| \kappa _j^{f^{hands}} - \hat{\kappa} _j^{f^{hands}} \|}} $, ${d}_{face} = \sum\limits_{k = 0}^{K} {\sqrt {\| \kappa _k^{f^{face}} - \hat{\kappa} _k^{f^{face}} \|}}$, and ${d}_{body} = \sum\limits_{i = 0}^{I} \sqrt {\| \kappa _i^{f^{body}} - \hat{\kappa} _i^{f^{body}} \|}$ are sum of Euclidean distance for hands, face, and body, $\kappa$ and $\hat{\kappa}$ are joints from DWPose on groundtruth and generated images, $I$, $J$, and $K$, are the total number of keypoints obtained from DWPose for the body, hands, and face regions, respectively, and ${\beta_{hands}}$, ${\beta_{face}}$, ${\beta_{body}}$ are weights for each loss contribution with ${\beta_{hands}} + {\beta_{face}} + {\beta_{body}} = 1$.

Finally, the RACL is used as a weight to scale the original diffusion-based MSE loss ${{\cal L}_\varepsilon }$ from Eq. \ref{eq:2}. This is because RACL will be larger and the scaled loss will penalize when the model generates inaccurate hand gestures, which helps the model to learn samples with complex hand features. The total loss is as follows:
 \begin{equation}
{{\cal L}_{{\rm{total}}}} = {{\cal L}_{RACL}} \cdot {{\cal L}_\varepsilon }.
  \label{eq:4}
\end{equation}

\section{Experiments}
\label{sec:blind}
\subsection{Implementation Details}
Our experiments are performed using two baseline models: the frozen Stable Diffusion XL model\footnote {\url{https://huggingface.co/docs/diffusers/using-diffusers/sdxl}} and the trainable ControlNet model\footnote {\url{https://huggingface.co/docs/diffusers/using-diffusers/ControlNet}}. Since ControlNet can also support multiple control input, we trained multiple ControlNets on different modalities, and then weighted sum the output embedding of these ControlNets as baseline fusion modalities. Although our proposed adaptive fusion mechanism can easily adjust the number of input modes, in following experiments, we used both DWPoseonDepth and surface normal modalities as inputs to the baseline ControlNet fusion and the proposed adaptive fusion mechanism, and set the weights of each modality in ControlNet fusion to be the same. We trained the ControlNet model for $90k$ steps on 83,864 curated frames from the How2Sign dataset, and evaluated the model with 15,637 annotated frames. We filtered these frames with a threshold value $\tau = 133$ and a hand keypoint confidence threshold of 0.2.
Our model can be trained and tested by diverse character data, but due to the limitation of computational resources we only trained and tested the filtered signer 5 data from the How2Sign dataset in following experiments. The resolution of the images is $512\times 512$. Since the How2Sign dataset does not contain text annotations for video content and text control is not the focus of this work, we used simple spaces, \ie `` '', for the text annotation accompanying the frames. 
Based on empirical evidence, we used ${\beta _{hands}} = 0.4$, ${\beta _{face}} = 0.2$, and ${\beta _{body}} = 0.4$ as weights for RACL. At inference, a DDIM sampler~\cite{song2020denoising} 
was used for image generation. All training was implemented on two NVIDIA Tesla 80GB-A100 GPUs.

\subsection{Quantitative results}
For the accurate comparison of the generation quality and pose alignment of hand regions in generated digital humans, we use two specific evaluation metrics for the hand region: \emph{hand-PSNR} and \emph{hand-Distance}. Specifically, 
hand-PSNR is computed by cropping and using only the hand region of the generated images and the ground truth images. Higher hand-PSNR indicates the better quality of the generated hand region. 
To compute hand-Distance, we utilise the DWPose annotator ~\cite{yang2023effective} to obtain the hand keypoint coordinates of the generated images, which is then used to get the Euclidean distance from ground truth keypoints. Lower hand-Distance indicates better generated hand accuracy of the corresponding comparison method.

The quantitative results are presented in Table \ref{tab:table1} and \ref{tab:table2}. As can be seen from these tables, the surface normal outperforms other methods as single control input, which suggests that the ControlNet models trained with the surface normal map are capable of obtaining richer pose information. However, all multi-modal fusion inputs outperform any single control modality, demonstrating that multiple modalities enable the model to learn richer posture features. 
Compared to ControlNet fusion, our proposed adaptive fusion with RACL has advantages in PSNR-related metrics and significantly improves performance in \emph{hand-Distance} while maintaining competitive performance of whole-body distance, indicating that the proposed adaptive fusion network and RACL are able to combine the advantages of different control modalities and allow the model to learn better pose features to further improve the performance of the hand region without sacrificing overall performance. Furthermore, our method has much lower computational complexity than ControlNet fusion, which requires training multiple controlnets on different modalities and loading them in the inference stage. In contrast, our method trains lightweight encoders for corresponding modalities and a weight prediction module. Interestingly, by comparing the performance of DWPose, depth, and DWPoseonDepth in terms of PSNR and distance, it can be found that depth outperforms DWPose in terms of PSNR-related performance, but underperforms DWPose in terms of distance-related performance. This is because the depth information for the whole image is not as good as DWPose in terms of pose control at the pixel level, especially when the hands are close to the chest. DWPoseonDepth performs particularly poorly in terms of pixel-level distance, due to the fact that the two modalities, DWPose and depth, may have conflicting annotations in the hand region (as shown in the yellow boxes in Fig. 1), increasing the complexity of learning from the control modalities.

\begin{table*}[t!]
\caption{Quantitative results on PSNR-related evaluation metrics produced with different conditional control modalities and losses.  It can be seen that our proposed adaptive multi-modal control fusion method with RACL produces the best results in hand-PSNR and whole-image-PSNR. Best quality is highlighted in \textbf{bold} and the second ranked method is \underline{underlined}.}\label{tab:table1}
\centering
\begin{tabular}{|l|c|c|}
\hline
Method  & hand-PSNR (dB) ↑ & whole-PSNR (dB) ↑ \\
\hline\hline
    DWPose & 15.77 & 19.90\\
    OpenPose & 16.08 & 19.82\\
    DWPoseonDepth & 16.34 & 20.79\\
    Depth & 17.15 & 20.92\\
    Normal & 19.70 & 22.99 \\
    ControlNet Fusion & \underline{19.59} & \underline{23.21} \\
    Adaptive Fusion + RACL (Ours) & \textbf{20.18} & \textbf{23.22}\\
\hline
\end{tabular}
\end{table*}

\begin{table*}[t!]
\caption{Quantitative results on distance-related evaluation metrics produced with different conditional control modalities and losses. Our proposed adaptive multi-modal control fusion method  with RACL produces the best results in hand-Distance while maintaining competitive whole-Distance performance. Best accuracy is highlighted in \textbf{bold} and the second ranked method is \underline{underlined}.}\label{tab:table2}
\centering
\begin{tabular}{|l|c|c|}
\hline
Method  & hand-Distance (pixels) ↓ & whole-Distance (pixels) ↓\\
\hline\hline
    DWPoseonDepth & 20.83 & 47.15\\
    OpenPose & 19.99 & 48.38\\
    Depth & 19.70 & 48.38\\
    DWPose & 19.35 & 47.13\\
    Normal & 15.62 & 35.89\\
    ControlNet Fusion & \underline{13.57} & \textbf{33.33} \\
    Adaptive Fusion + RACL (Ours) & \textbf{11.72} & \underline{33.93}\\
\hline
\end{tabular}
\end{table*}

\begin{figure}[!htb]
  \centering
  \includegraphics[height=12cm]{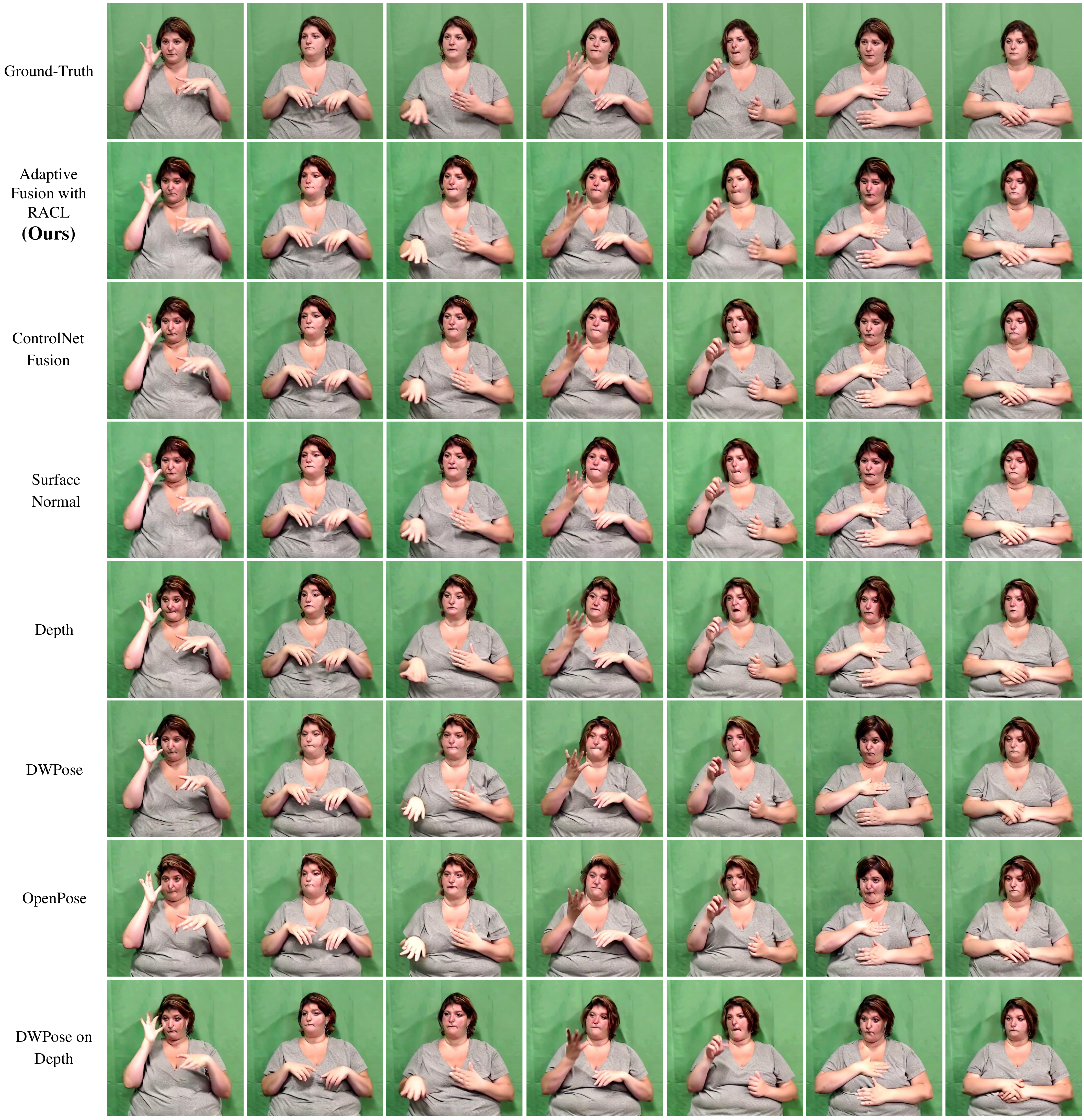}
      \caption{Qualitative performance comparison of the proposed method with different methods for conditional control. Gradual deterioration of qualitative performance can be observed from top to bottom row. It can be seen that the proposed adaptive multi-modal control fusion with RACL performs better hand gestures control and image quality than other methods. Please zoom in for details.
  }
  \label{fig:Fig5}
\end{figure}

\subsection{Qualitative results}
Figure \ref{fig:Fig5} illustrates the qualitative comparison of poses generated by the comparison method using different combinations of modalities. It can be seen that for a single control input, the surface normal mode performs best, but fails to generate accurate fingers when motion blur occurs. This problem is significantly improved when multi-modal control information is fused using our proposed method. 
This is because the proposed adaptive fusion module can compensate for the missing information in one modality through the others. Similarly, it can be observed that drawing the DWPose skeleton on the depth map as an integral input is worse than the DWPose skeleton alone or the depth map alone as a control input. Specifically, the hand generated using DWPoseonDepth shows more blur, and the fingers appear difficult to distinguish compared to using DWPose skeleton or depth as input. 
This is because DWPoseonDepth highlights the conflicts between DWPose and depth modalities in the detailed annotation of the hand, making it more difficult for the model to learn accurate pose information.

\subsection{Ablation Studies}

Table \ref{tab:table3} and \ref{tab:table4} show the ablation experiments of the proposed method. We tested the results for the proposed adaptive multi-control fusion input and surface normal input with (w) and without (wo) RACL. It can be seen that the proposed multi-control input exceeds the Normal input in all performance. This indicates that the proposed adaptive fusion module is able to get better fusion control information from the multi-modal inputs and improve the quality of human pose generation. Moreover, by comparing the results of different inputs with and without RACL, it can be seen that the RACL improves hand-Distance performance without much sacrifice of other performance, which indicates that RACL provides more accurate control of the hand gestures while maintaining the accuracy of the whole body posture.

\begin{table*}[t!]
\caption{Ablation studies on PSNR-related evaluation metrics produced with different conditional control modalities and losses.  It can be seen that our proposed adaptive multi-modal control fusion method produces better results in hand-PSNR and whole-image-PSNR. However, the proposed RACL does not steadily improve the performance of the evaluation metrics related to PSNR. Best quality is highlighted in \textbf{bold} and the second ranked method is \underline{underlined}.}\label{tab:table3}
\centering
\begin{tabular}{|l|c|c|}
\hline
Method  & hand-PSNR (dB) ↑ & whole-PSNR (dB) ↑ \\
\hline\hline
    Normal & 19.70 & 22.99 \\
    Normal + RACL & 19.81 & \underline{23.25} \\
    Adaptive Fusion + RACL & \underline{20.18} & 23.21\\
    Adaptive Fusion & \textbf{20.40} & \textbf{23.30}\\
\hline
\end{tabular}
\end{table*}

\begin{table*}[t!]
\caption{Ablation studies on distance-related evaluation metrics produced with different conditional control modalities and losses. 
Our proposed adaptive multi-modal control fusion method shows advantages in hand-Distance and whole-body-Distance. Furthermore, the proposed RACL also improves the performance of the hand distance while keeping whole-body-Distance nearly unchanged. The best accuracy is highlighted in \textbf{bold} and the second-ranked method is \underline{underlined}.}\label{tab:table4}
\centering
\begin{tabular}{|l|c|c|}
\hline
Method  & hand-Distance (pixels) ↓ & whole-Distance (pixels) ↓\\
\hline\hline
    Normal & 15.62 & 35.89\\
    Normal + RACL & 14.53 & 35.19 \\
    Adaptive Fusion & \underline{12.88} & \textbf{33.77}\\
    Adaptive Fusion + RACL & \textbf{11.72} & \underline{33.93}\\
\hline
\end{tabular}
\end{table*}

\begin{figure}[!htb]
  \centering
  \includegraphics[height=4.8cm]{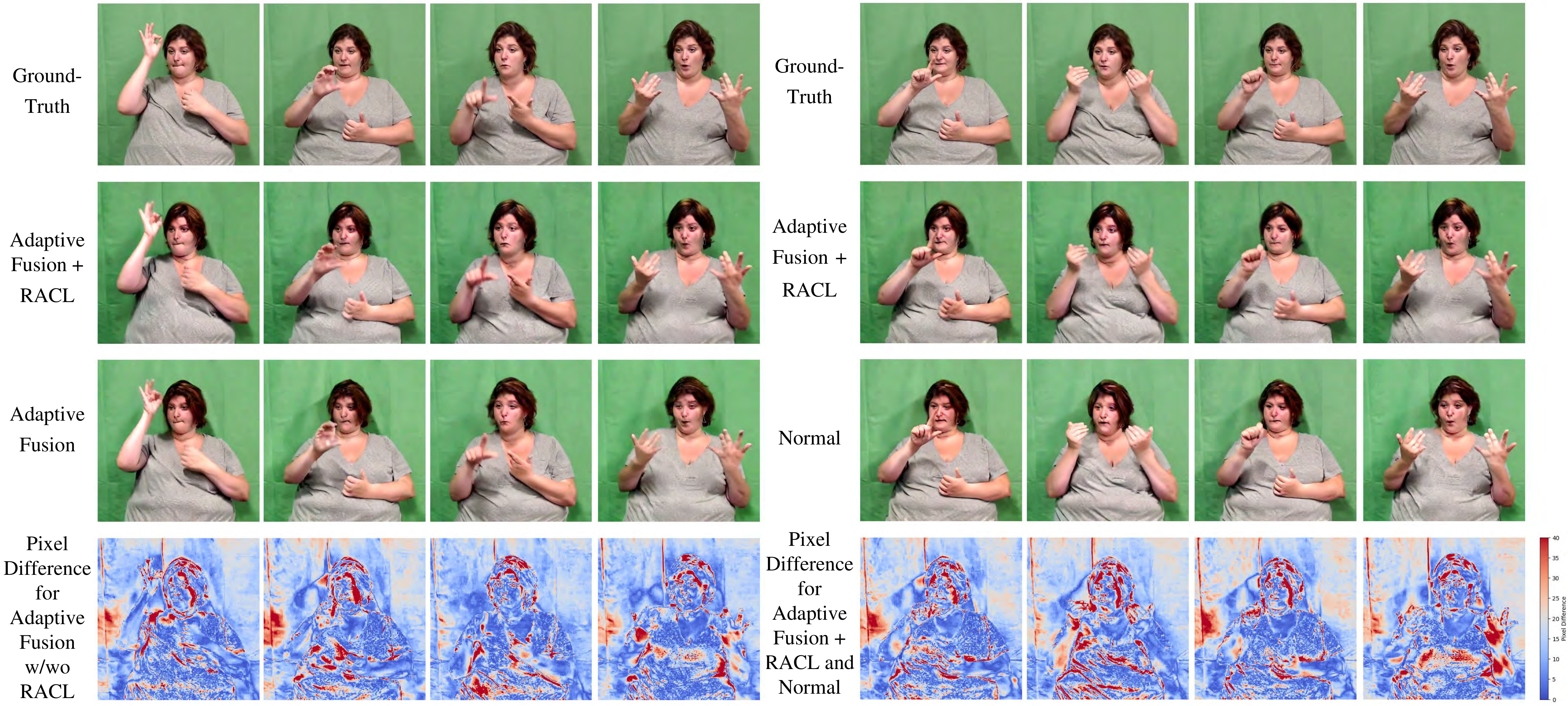}
      \caption{Qualitative performance ablation studies of surface normal input and the proposed adaptive multi-modal fusion method with (w) or without (wo) RACL. It can be seen that RACL has impact on the skeletal information of the hand and face. The multimodal fusion module also has an effect on background generation.
  }
  \label{fig:Fig6}
\end{figure}

Figure \ref{fig:Fig6} shows qualitative performance comparison of surface normal input and the proposed adaptive multi-modal fusion method with or without RACL. In the comparison of different methods, we show samples with the largest improvement in accuracy when using RACL in the fourth line, where we visualise pixel differences as heat maps. The heat map in the four columns on the left shows the effect of RACL on multi-modal inputs. As can be seen from the heat map in the four left columns, effects of using RACL on both multi-modal inputs are mainly concentrated on the human body, with the larger differences especially in the areas related to the hands, arms, and face. The heat map in the four columns on the right shows the difference between the proposed method and a single surface normal input. There is a clear difference in the background in addition to the large difference in the human hand and face. This is because the proposed adaptive multi-modal fusion method affects the entire image.

\section{Conclusion}
This paper proposed a novel pipeline for improving hand pose accuracy for digital humans generation. To achieve this, we first designed a data pre-processing pipeline for cleaning and relabelling the How2Sign dataset. Then, an adaptive multi-modal control fusion network was designed for feature fusion of different control modal inputs to train a ControlNet-based Diffusion model. To further improve the accuracy of generated hand poses while maintain the whole body pose accuracy, we designed a Region-Aware Cycle Loss (RACL) to train the model. Meanwhile, we used two evaluation metrics for quality and hand pose consistency in the hand region. Experimental results demonstrate the effectiveness of the proposed adaptive multi-modal control fusion network and RACL.

\section*{Acknowledgements}
This research was supported by the China Scholarship Council. We utilized Queen Mary University of London's Andrena HPC facility, and acknowledge the assistance of the ITS Research team.

%
%
\bibliographystyle{splncs04}
\bibliography{main}

\end{document}